\PassOptionsToPackage{hyphens}{url}
\PassOptionsToPackage{numbers,sort&compress}{natbib}
\RequirePackage[svgnames,table]{xcolor}
\documentclass[11pt,letterpaper,logo]{yalearxiv}

\usepackage{natbib}
\usepackage{multirow,array,adjustbox}
\usepackage{flafter,float}
\definecolor{EstimateColor}{RGB}{0,0,0}
\definecolor{MutedBlue}{RGB}{36,65,95}
\newcommand{\rnum}[1]{\textcolor{EstimateColor}{#1}}
\newcommand{\rtext}[1]{\textcolor{EstimateColor}{#1}}

\newcommand{\tailacc}{\operatorname{Acc}_{\mathrm{tail}}}

\newcommand{\pp}{~pp}

\newcommand{\ControlledFiveCEGain}{0.17}
\newcommand{\ControlledFiveBCLGain}{0.26}

\newcommand{\ControlledFiveGeoGain}{0.28}
\newcommand{\ControlledFiveGeoUCB}{0.45}
\newcommand{\CoupledFiveCEGain}{1.66}
\newcommand{\CoupledFiveBCLGain}{1.92}

\newcommand{\DecisionChangeBCL}{1.66}
\newcommand{\DecisionChangeBCLLow}{1.62}
\newcommand{\DecisionChangeBCLHigh}{1.70}
\newcommand{\AssignmentClusterChangeLCB}{1.57}
\newcommand{\SupportAccuracyDrop}{21.30}
\newcommand{\BalancedSupportMaxUCB}{0.45}
\newcommand{\StageOneBalancedPostSkewGain}{0.34}
\newcommand{\StageOneSkewPostBalancedGain}{1.86}
\newcommand{\ExtraTailDrawsGain}{0.43}
\newcommand{\GradientWeightGain}{0.51}
\newcommand{\PathMaxDifference}{0.18}
\newcommand{\ImageNetFiveBalancedGain}{0.34}
\newcommand{\ImageNetFiveBalancedUCB}{0.56}
\newcommand{\ImageNetFiveSkewGain}{1.70}
\newcommand{\ImageNetFiveSkewLCB}{1.42}
\newcommand{\ImageNetLTOverallGain}{0.80}
\newcommand{\ImageNetLTFewGain}{1.60}
\newcommand{\ImageNetLTOverallLow}{0.55}
\newcommand{\ImageNetLTOverallHigh}{1.05}
\newcommand{\ImageNetLTFewLow}{1.19}
\newcommand{\ImageNetLTFewHigh}{2.01}
\newcommand{\ImageNetLTRepresentationRuns}{3}

\newcommand{\RobustBalancedMaxUCB}{0.61}
\newcommand{\RobustSkewMinLCB}{1.25}

\newcommand{\CEBalancedGain}{0.38}
\newcommand{\CEBalancedUCB}{0.61}
\newcommand{\CESkewGain}{1.56}
\newcommand{\CESkewLCB}{1.25}
\newcommand{\GeometrySpearman}{0.94}
\newcommand{\BalancedMinSupportedMargin}{0.45}
\newcommand{\CoupledMaxSupportedMargin}{1.45}

\title{Beyond Scarcity: When Encoder Adaptation Is Worthwhile}
\runningtitle{Beyond Scarcity: When Encoder Adaptation Is Worthwhile}
\author{Zexing Zhang, Jichao Li\textsuperscript{\textdagger}, Tianyang Lei, Yiyin Zhang, and Kewei Yang\\
{\normalfont College of Systems Engineering, National University of Defense Technology}\\
{\normalfont Changsha 410073, China}\\
{\normalfont\textsuperscript{\textdagger} Corresponding author}}
\hypersetup{
  colorlinks=true,
  linkcolor=blue!50!black,
  citecolor=blue!50!black,
  urlcolor=blue!50!black,
  pdftitle={Beyond Scarcity: When Encoder Adaptation Is Worthwhile},
  pdfauthor={Zexing Zhang, Jichao Li, Tianyang Lei, Yiyin Zhang, Kewei Yang}
}

\begin{document}

\begin{abstract}
Long-tailed recognition often learns a shared representation before correcting the classifier for class imbalance. Once the classifier has been rebalanced, however, final accuracy does not reveal whether further encoder adaptation remains worthwhile. Standard imbalance benchmarks obscure this question because they jointly vary the number of distinct examples available to rare classes and the frequency with which those examples shape the representation. We introduce a controlled evaluation framework that separates unique support from representation allocation and estimates the incremental value of encoder updates through matched classifier-only and full-network continuations from a common corrected model. The framework combines stage-specific controls, multiple representation and adaptation objectives, whole-distribution metrics, and hierarchical inference over class assignments, support reconstructions, and optimization runs. \rtext{Across the controlled study, the same low-support regime moves from one in which classifier-only training is adequate to one in which encoder adaptation yields a practically meaningful benefit solely as representation allocation becomes more imbalanced. The transition persists across classifier corrections, representation objectives, normalization and augmentation controls, and ImageNet-scale evaluation, while preserving frequent-class accuracy. Under the prespecified one-percentage-point operating margin, the two regimes are separated by simultaneous confidence bounds.} These results show that benchmark construction can alter a post-correction training decision and that support size alone does not characterize the remaining value of representation adaptation.
\end{abstract}

\maketitle

\section{Introduction}

A widely used strategy for long-tailed recognition is to learn a shared representation and then correct the classifier for class imbalance. Class-balanced retraining can recover substantial rare-class accuracy while leaving the encoder fixed \citep{kang2020decoupling,sun2025los}, and recent lightweight or classifier-focused adaptation methods make this alternative increasingly competitive \citep{shi2024lift,zhao2025tsmof}. This efficiency creates an important stopping decision. After strong classifier correction, is the representation already adequate for the available data, or is further full-network adaptation worth its additional computation?

The answer is usually inferred from comparisons among complete training pipelines. Such comparisons are useful for ranking methods, but they combine differences in representation learning, classifier fitting, and optimization history. Their final accuracy gap therefore does not measure the accuracy that remains available from updating the encoder after classifier correction. A direct answer requires the classifier-only and full-network alternatives to begin from the same corrected model and to differ only in whether the encoder receives gradients.

Benchmark construction introduces a second ambiguity. Synthetic long-tailed benchmarks commonly increase severity by reducing the number of examples in low-frequency classes. The same construction also reduces how often those classes participate while the representation is learned. Severity consequently changes both the evidence available to a rare class and the influence of that evidence on optimization. These factors need not imply the same remedy. Additional unique examples expand the empirical support, whereas repeated class exposure changes how the existing examples shape the representation.

\begin{figure}[!htbp]
  \centering
  \includegraphics[width=\textwidth]{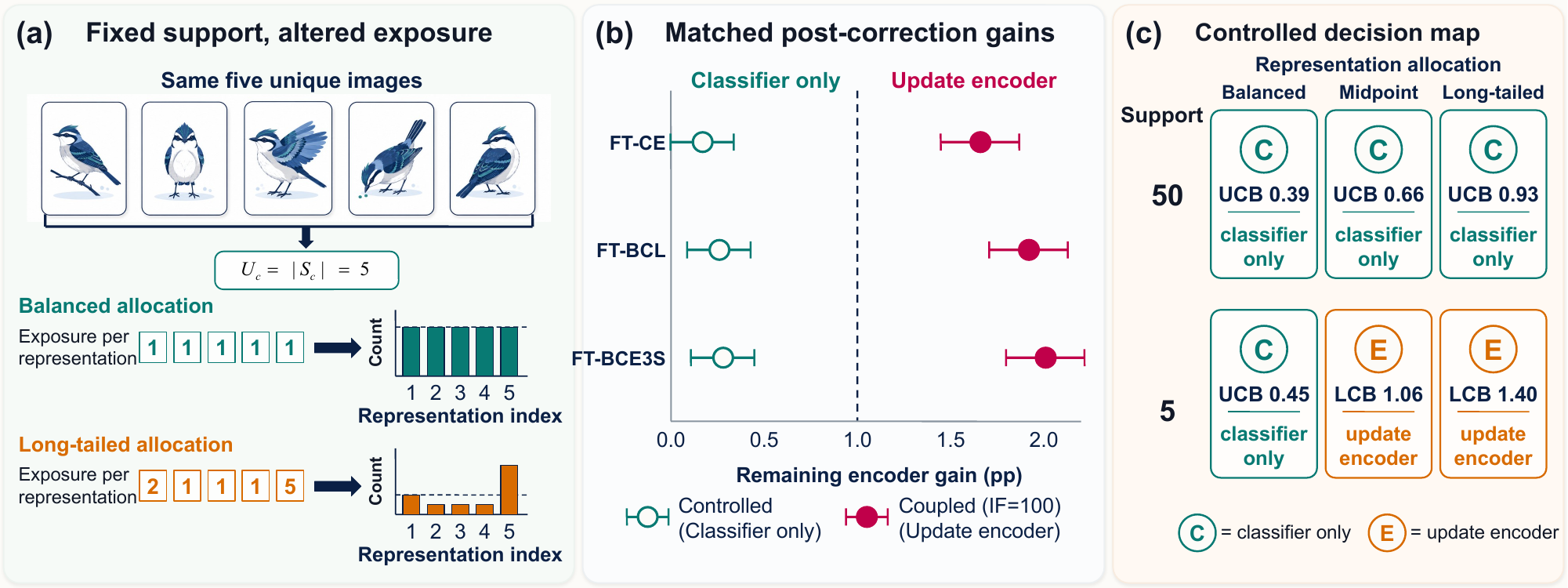}
  \caption{Benchmark construction changes the post-correction training choice. (a) The available tail images are fixed while their representation allocation changes. (b) \rtext{The coupled benchmark and controlled profile place every matched adaptation objective on opposite sides of the prespecified operating margin.} (c) The displayed support-by-allocation grid reports the decision-relevant family bound for the two endpoint support levels and three allocation regimes. C denotes classifier-only training and E denotes encoder adaptation.}
  \label{fig:teaser}
\end{figure}

Consider a rare class in an extreme low-support regime. Two training schedules may expose the learner to exactly the same source images but allocate very different numbers of representation-learning updates to that class. The information available from the dataset is unchanged, yet the learned representation can reflect that information to different degrees. This distinction suggests a sharper question than whether lower support makes recognition harder: does benchmark coupling change the remaining value of encoder adaptation after classifier correction?

We address this question with a controlled factorization of long-tail severity. \emph{Unique support} records the number of distinct source images available to a class. \emph{Representation allocation} records the class-level draw budget used while learning the shared representation. We vary these quantities separately while holding the architecture, available images, augmentation family, and optimization budget fixed. For every controlled condition, a common corrected model is continued along two matched branches. One branch updates only the classifier and the other updates the full network. The paired difference estimates the incremental value of encoder adaptation under the same data stream, views, training duration, and classifier schedule. Figure~\ref{fig:teaser} summarizes the central comparison.

\rtext{The controlled comparison reveals a consistent transition. In the lowest-support condition, balanced representation allocation leaves all three adaptation objectives below the practical margin, whereas long-tailed allocation moves all three beyond it. The ordering survives controls over class mixture, normalization, augmentation reuse, classifier correction, and the representation objective. A pre-adaptation feature-geometry diagnostic changes with the same allocation history, and controlled ImageNet-100 reproduces the transition. Repeated ImageNet-LT runs show a positive encoder benefit under natural imbalance.}

This work makes three contributions.
\begin{itemize}
  \item We formulate long-tail severity as the combination of unique support and representation allocation, and introduce a stage-explicit design that varies these factors without changing the available images or the later comparison.
  \item We develop a matched decision framework that measures residual encoder value after classifier correction, localizes the effect to representation learning, and quantifies uncertainty across class assignments, support reconstructions, and optimization runs.
  \item \rtext{We show that the coupling used to construct standard long-tail benchmarks can change the preferred post-correction training strategy at fixed support. The result persists across stronger correction methods, representation and class-mixture controls, whole-distribution metrics, practical margins, and ImageNet-scale evaluation.}
\end{itemize}

\begin{figure}[!htbp]
  \centering
  \includegraphics[width=\textwidth]{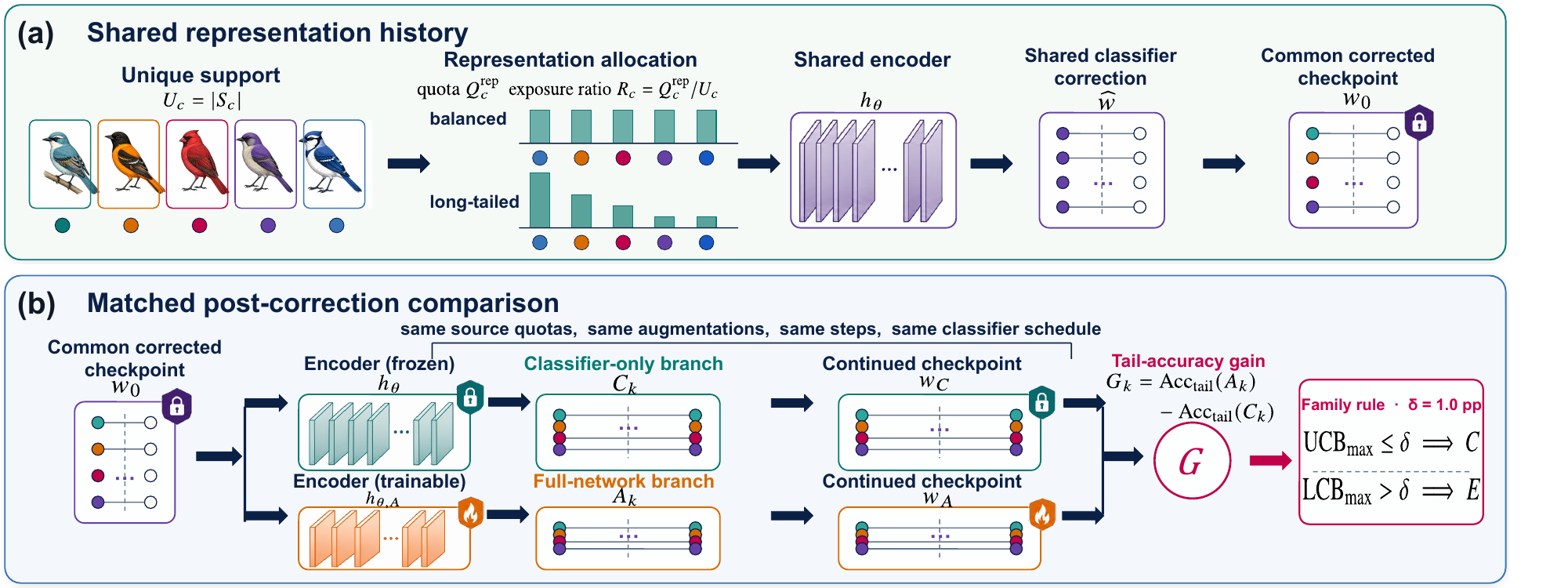}
  \caption{Controlled profiles and matched post-correction comparison. Unique support specifies the images available to each class. Representation allocation controls their class-level exposure while the shared encoder is learned. Classifier correction then produces a common model state. The two later branches receive identical inputs and schedules and differ only in whether encoder parameters are updated. Here, $\mathrm{UCB}_{\max}=\max_k\mathrm{UCB}(G_k)$ and $\mathrm{LCB}_{\max}=\max_k\mathrm{LCB}(G_k)$ summarize the familywise stopping rule at $\delta=1.0$~pp.}
  \label{fig:method}
\end{figure}


\section{Related Work}

\paragraph{Staged correction and efficient adaptation.}
Long-tailed recognition methods intervene during representation learning, classifier fitting, or inference \citep{fu2022survey}. Decoupled training showed that rebalancing a classifier on a fixed representation can recover substantial tail accuracy \citep{kang2020decoupling}. Subsequent work has strengthened this stage through prior correction, calibration, target smoothing, and classifier-geometry control \citep{ren2020balanced,menon2021logit,hong2021lade,zhong2021mislas,sun2025los,zhang2026samn,terven2026bscrt}. Efficient adaptation has also become central. LIFT finds that heavy tuning of foundation models can damage tail performance and develops a lightweight alternative \citep{shi2024lift}, while TS-MOF freezes the feature extractor and optimizes specialized classifier heads in a second stage \citep{zhao2025tsmof}. These studies establish strong low-cost alternatives to full-network training. Our focus is complementary: after a fixed correction has been applied, we estimate the incremental value of encoder updates under matched later training and ask whether benchmark construction changes that value.

\paragraph{Long-tailed representation learning.}
Long-tailed representations have been improved through complementary experts, semantic augmentation, contrastive objectives, and joint feature--classifier geometry \citep{zhou2020bbn,cai2021ace,li2021metasaug,cui2021paco,zhu2022bcl,li2022gcl,fan2026bce3s}. These methods demonstrate that representation quality matters, but their final pipeline comparisons combine the earlier representation, the classifier, and the subsequent optimization path. They therefore do not identify how much benefit remains from updating an encoder once classifier optimization is held fixed. The matched continuations in our study turn that residual benefit into a paired quantity, and a matched-forward-view cross-entropy control tests whether the observed transition is specific to contrastive representation learning.

\paragraph{Sampling and benchmark construction.}
Resampling changes both the class mixture seen by optimization and the reuse of scarce examples. Breadcrumbs analyzes overfitting under repeated class-balanced sampling \citep{liu2021breadcrumbs}, and retrieval-augmented classification offers an alternative way to exploit limited tail evidence \citep{long2022rac}. Standard severity constructions additionally reduce the set of distinct rare-class examples while changing their training exposure. Our factorization separates these two properties. This reveals a consequence not visible from final accuracy alone: at fixed support, the coupling built into a benchmark can move the post-correction encoder gain across a practical decision boundary.


\section{Method}

We seek the accuracy attributable to updating an encoder after strong classifier correction. Figure~\ref{fig:method} separates the earlier representation history from the later choice between classifier-only and full-network training.

\subsection{Support and Representation Allocation}

For class $c$, let $S_c$ be the set of distinct source images available during training and let $U_c=|S_c|$ denote its unique support. Representation learning assigns the class an exact source-draw quota $Q_c^{\mathrm{rep}}$. Sampling is uniform with replacement within $S_c$, so the average exposure per available image is
\begin{equation}
  R_c=\frac{Q_c^{\mathrm{rep}}}{U_c}.
  \label{eq:reuse}
\end{equation}
Thus, a class with ten images and thirty source draws has $R_c=3$, whereas assigning three draws gives $R_c=0.3$ without changing which images are available. $U_c$ controls the empirical evidence that can be observed. $Q_c^{\mathrm{rep}}$ controls how strongly that evidence participates in representation learning.

Head and middle support remain fixed and tail support varies through nested subsets. For a fixed subset, quotas interpolate between class-balanced allocation and the role-level allocation induced by a severe long-tailed profile. Integer quotas are obtained by largest-remainder rounding, which preserves the total number of source draws exactly. Three coarse-stratified class assignments rotate semantic groups through head, middle, and tail roles.

\subsection{Stage-Specific Comparison}

The controlled allocation is used only to learn the shared representation. Classifier correction and both later branches use balanced quotas. This design tests whether representation history changes the value of a subsequent encoder update while keeping the later sampling schedule fixed. Table~\ref{tab:main-stages} makes the stage boundary explicit, and a cross-stage experiment varies the two schedules independently.

\begin{table}[!htbp]
\centering
\small
\setlength{\tabcolsep}{7.0pt}
\begin{adjustbox}{max width=\linewidth}
\begin{tabular}{llll}
\toprule
Stage & Source allocation & Trainable parameters & Role in the comparison \\
\midrule

Representation learning & controlled profile & encoder and classifier & learns the shared representation \\
Classifier correction & class-balanced & classifier only & produces the common corrected model \\
Classifier continuation & class-balanced & classifier only & controls additional classifier optimization \\
Full-network adaptation & class-balanced & encoder and classifier & measures the remaining encoder gain \\


\bottomrule
\end{tabular}
\end{adjustbox}
\caption{Stage-wise training design. The controlled representation allocation is confined to representation learning. The matched post-correction branches use the same balanced source quotas and differ only in whether the encoder is trainable.}
\label{tab:main-stages}
\end{table}

\subsection{Matched Encoder Gain}

A classifier correction method is selected once on a disjoint development profile and fixed thereafter. For adaptation objective $k$, the corrected model is continued along a classifier-only branch $C_k$ and a full-network branch $A_k$. The branches share source indices, augmentation views, update count, classifier learning-rate schedule, and evaluation time. Their difference
\begin{equation}
  G_k=\tailacc(A_k)-\tailacc(C_k)
  \label{eq:gain}
\end{equation}
removes the classifier optimization common to both branches. $G_k$ is an operational estimate for the specified objective and budget, not a path-independent decomposition of classifier and representation effects. We evaluate cross-entropy, BCL, and a BCE3S-style adaptation objective \citep{fan2026bce3s}.

\subsection{Decision Rule and Uncertainty}

Accuracy levels are reported in percent and accuracy differences in percentage points, abbreviated as pp. The primary operating margin is one percentage point in tail macro top-1 accuracy. Classifier-only training is considered adequate relative to the tested family when the simultaneous upper bounds for all $G_k$ values lie below this margin. Encoder adaptation is selected when at least one simultaneous lower bound exceeds it. The bounds and the complete margin curve are reported so that other cost tolerances can be applied.

Each principal CIFAR condition contains three class-role assignments, four support reconstructions per assignment, and two optimization runs per profile. The primary interval uses a hierarchical bootstrap over these levels. Paired intervals over profile means, leave-one-assignment-out estimates, and assignment-level clustering provide sensitivity checks. The target is variation over support reconstruction and optimization for the fixed CIFAR-100 class universe.

\FloatBarrier


\section{Experiments}

\subsection{Experimental Setup}

\paragraph{Datasets and profiles.}
CIFAR-100 experiments use ResNet-32. Head classes retain 500 images, middle classes retain 100, and tail support takes four nested levels from 50 to 5 images. The controlled grid combines support with balanced, midpoint, and long-tailed representation allocation. Coupled references use the conventional exponential profiles with imbalance factors 10, 20, 50, and 100. Controlled ImageNet-100 uses ResNet-50, two class-role assignments, three support reconstructions per assignment, and two support levels. ImageNet-LT provides the natural-benchmark evaluation \citep{liu2019oltr}.

\paragraph{Methods and metrics.}
BCL supplies the primary shared representation \citep{zhu2022bcl}; a matched-forward-view CE representation tests dependence on contrastive learning. The classifier side includes cRT \citep{kang2020decoupling}, LOS \citep{sun2025los}, SAMN \citep{zhang2026samn}, and BS-cRT \citep{terven2026bscrt}. LOS is selected once on a disjoint development profile for the primary comparison. The matched adaptation family contains FT-CE, FT-BCL, and FT-BCE3S. BCE3S is also evaluated with its published training pipeline \citep{fan2026bce3s}. The primary metric is tail macro top-1 accuracy. Head, middle, overall, balanced, and tail lower-quartile accuracy provide distribution-wide checks, and GPU-hours quantify additional computation.

\subsection{Benchmark Construction Changes the Training Choice}

Figure~\ref{fig:teaser} compares the coupled IF=100 profile with a controlled condition at the same lowest support. \rtext{Under the coupled profile, FT-CE, FT-BCL, and FT-BCE3S gain \rnum{\CoupledFiveCEGain}, \rnum{\CoupledFiveBCLGain}, and \rnum{2.01}\pp{} over their matched classifier continuations. Every simultaneous lower bound exceeds the one-point operating margin. With balanced representation allocation, the corresponding gains are \rnum{\ControlledFiveCEGain}, \rnum{\ControlledFiveBCLGain}, and \rnum{\ControlledFiveGeoGain}\pp{}, and every simultaneous upper bound is at most \rnum{\ControlledFiveGeoUCB}\pp.} Table~\ref{tab:decision} reports the complete family comparison.

\begin{table}[!htbp]
\centering
\scriptsize
\setlength{\tabcolsep}{3.6pt}
\begin{adjustbox}{max width=\linewidth}
\begin{tabular}{lllcccll}
\toprule
Dataset & Profile & Representation allocation & FT-CE & FT-BCL & FT-BCE3S & Family bound & Decision \\
\midrule

CIFAR-100 & Coupled IF=100 & long-tailed & \rnum{1.66} & \rnum{1.92} & \rnum{2.01} & all LCB $\geq$ \rnum{1.45} & \rtext{Update encoder} \\
CIFAR-100 & Controlled & balanced & \rnum{0.17} & \rnum{0.26} & \rnum{0.28} & all UCB $\leq$ \rnum{0.45} & \rtext{Classifier only} \\
CIFAR-100 & Controlled & midpoint & \rnum{1.25} & \rnum{1.31} & \rnum{1.38} & all LCB $\geq$ \rnum{1.06} & \rtext{Update encoder} \\
CIFAR-100 & Controlled & long-tailed & \rnum{1.61} & \rnum{1.86} & \rnum{1.94} & all LCB $\geq$ \rnum{1.40} & \rtext{Update encoder} \\
ImageNet-100 & Controlled & balanced & \rnum{0.24} & \rnum{0.34} & \rnum{0.38} & all UCB $\leq$ \rnum{0.60} & \rtext{Classifier only} \\
ImageNet-100 & Controlled & long-tailed & \rnum{1.43} & \rnum{1.70} & \rnum{1.78} & all LCB $\geq$ \rnum{1.17} & \rtext{Update encoder} \\


\bottomrule
\end{tabular}
\end{adjustbox}
\caption{Remaining encoder gain at the lowest tail-support level. Gains are in pp. The family-bound column reports the largest simultaneous upper confidence bound (UCB) when all objectives support classifier-only training, and the smallest simultaneous lower confidence bound (LCB) when encoder adaptation is selected.}
\label{tab:decision}
\end{table}

\rtext{For FT-BCL, benchmark coupling increases the remaining encoder gain by \rnum{\DecisionChangeBCL}\pp{}, with a 95\% interval of [\rnum{\DecisionChangeBCLLow}, \rnum{\DecisionChangeBCLHigh}]. The assignment-level analysis retains a lower bound of \rnum{\AssignmentClusterChangeLCB}\pp.} The decision change therefore survives both support reconstruction and a conservative treatment of the three semantic assignments.

\subsection{Support Controls Difficulty, Allocation Controls Residual Gain}

\begin{figure}[!htbp]
  \centering
  \includegraphics[width=\textwidth]{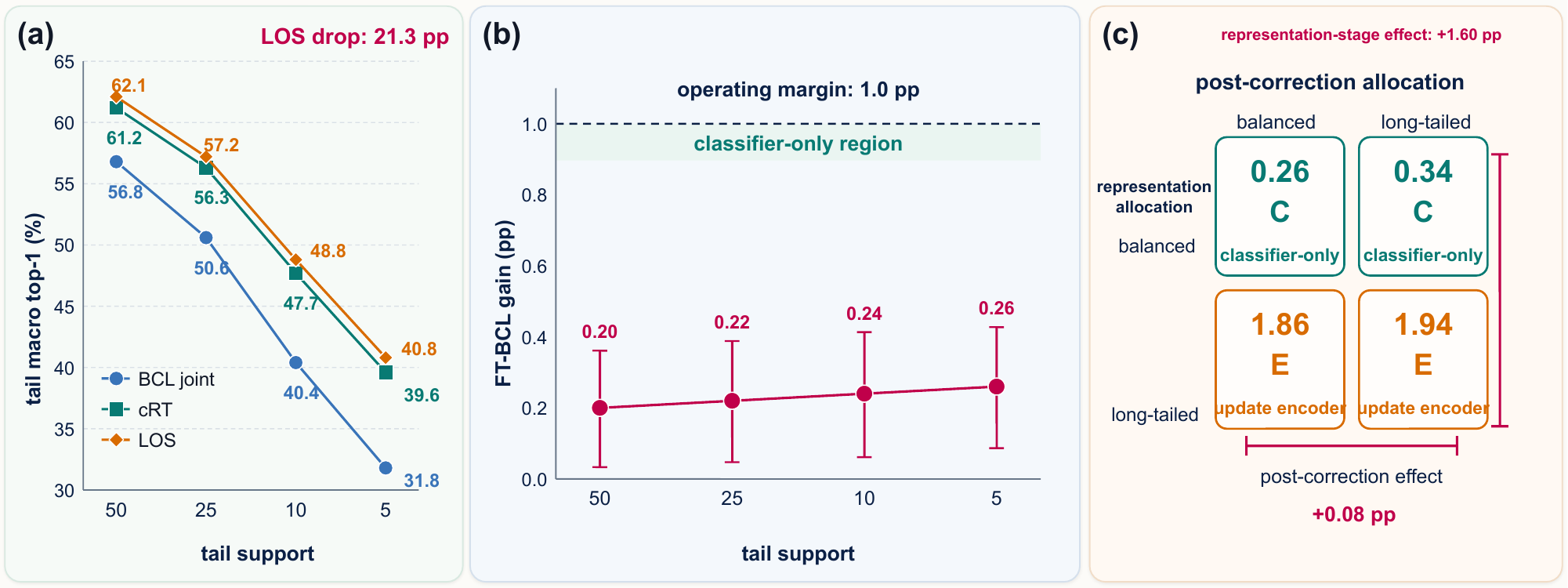}
  \caption{Support and allocation govern different outcomes. (a) \rtext{With balanced representation allocation, reducing support lowers tail accuracy by \rnum{\SupportAccuracyDrop}\pp.} (b) \rtext{The remaining FT-BCL gain stays below a simultaneous UCB of \rnum{\BalancedSupportMaxUCB}\pp{} over the same sweep.} (c) \rtext{Changing only the later allocation yields \rnum{\StageOneBalancedPostSkewGain}\pp{}, whereas changing the representation allocation alone yields \rnum{\StageOneSkewPostBalancedGain}\pp.}}
  \label{fig:factors}
\end{figure}

Figure~\ref{fig:factors}a holds representation allocation balanced while tail support decreases. \rtext{LOS accuracy falls from \rnum{62.1}\% at the highest support to \rnum{40.8}\% at the lowest, a decline of \rnum{\SupportAccuracyDrop}\pp.} Recognition becomes substantially harder as unique evidence disappears. Figure~\ref{fig:factors}b shows a different response for encoder adaptation. \rtext{Across the same support sweep, the FT-BCL gain remains between \rnum{0.20} and \rnum{0.26}\pp{}, and every simultaneous upper bound is at most \rnum{\BalancedSupportMaxUCB}\pp.} Absolute difficulty and the marginal value of updating the encoder are therefore distinct quantities.

The cross-stage experiment in Fig.~\ref{fig:factors}c locates the transition. \rtext{When representation learning is balanced, changing the later schedule from balanced to long-tailed raises the FT-BCL gain only from \rnum{0.26} to \rnum{0.34}\pp{}. When the representation is learned with long-tailed allocation, restoring balanced quotas for both later branches leaves a gain of \rnum{1.86}\pp{} with a lower bound of \rnum{1.58}\pp.} The representation-learning schedule, rather than the post-correction sampler, accounts for the substantive shift.

\subsection{Controls and Representation Evidence}

A fixed source budget couples greater tail exposure with fewer head and middle draws. Two controls test this alternative explanation. One adds tail draws without reducing frequent-class quotas. The other leaves the input mixture unchanged and increases only the tail gradient weight. \rtext{Their FT-BCL gains are \rnum{\ExtraTailDrawsGain} and \rnum{\GradientWeightGain}\pp{}, with upper bounds of \rnum{0.64} and \rnum{0.73}\pp.} Reduced frequent-class exposure is therefore not required to recover the classifier-only region.

\begin{figure}[!htbp]
  \centering
  \includegraphics[width=\textwidth]{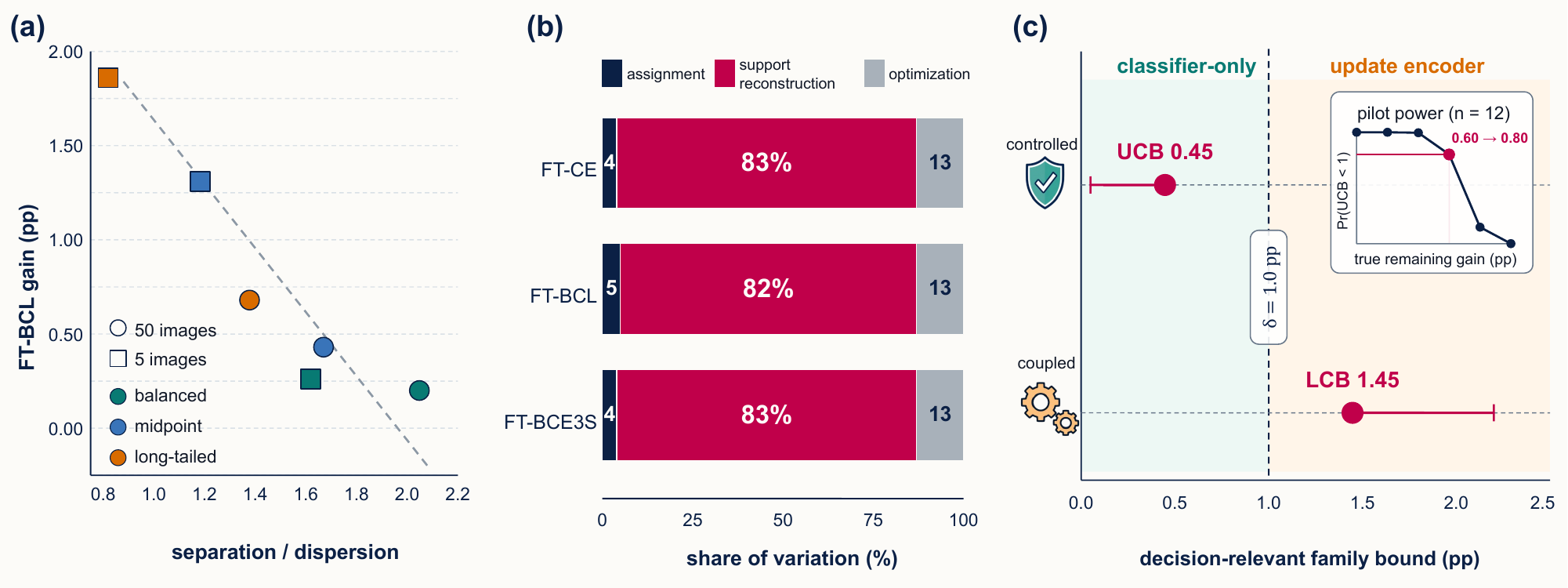}
  \caption{Representation evidence and decision reliability. (a) A pre-adaptation separation-to-dispersion ratio provides a descriptive correlate of the later FT-BCL gain. Circles and squares denote the two support levels; teal, blue, and amber denote balanced, midpoint, and long-tailed allocation. (b) \rtext{Variation is decomposed across class-role assignment, support reconstruction, and optimization.} (c) \rtext{The largest controlled UCB and smallest coupled LCB lie on opposite sides of the operating margin; the inset shows the pilot-based power curve for twelve profiles.}}
  \label{fig:robustness}
\end{figure}

Table~\ref{tab:matched-robustness} varies the corrected model, normalization policy, augmentation reuse, and representation objective while preserving the matched estimand. The first three rows begin from cRT-, LOS-, and SAMN-corrected models. The CE-3V row replaces the contrastive representation with three matched classification views.

\begin{table}[!htbp]
\centering
\scriptsize
\setlength{\tabcolsep}{3.0pt}
\begin{adjustbox}{max width=\linewidth}
\begin{tabular}{lcc}
\toprule
Correction or control & Balanced & Long-tailed \\
\midrule

cRT correction & \rnum{0.35} [\rnum{0.57}] & \rnum{1.89} [\rnum{1.60}] \\
LOS correction & \rnum{0.26} [\rnum{0.46}] & \rnum{1.86} [\rnum{1.58}] \\
SAMN correction & \rnum{0.30} [\rnum{0.49}] & \rnum{1.79} [\rnum{1.52}] \\
Frozen BN statistics & \rnum{0.29} [\rnum{0.49}] & \rnum{1.74} [\rnum{1.46}] \\
Recalibrated BN statistics & \rnum{0.27} [\rnum{0.47}] & \rnum{1.82} [\rnum{1.54}] \\
Fixed augmentation views & \rnum{0.32} [\rnum{0.53}] & \rnum{1.71} [\rnum{1.43}] \\
CE-3V representation & \rnum{0.38} [\rnum{0.61}] & \rnum{1.56} [\rnum{1.25}] \\


\bottomrule
\end{tabular}
\end{adjustbox}
\caption{Matched robustness at the lowest support. Each entry is mean FT-BCL gain followed by its decision-relevant bound in brackets. The balanced column gives the UCB; the long-tailed column gives the LCB. Values are in pp.}
\label{tab:matched-robustness}
\end{table}

\rtext{Every balanced-allocation upper bound in Table~\ref{tab:matched-robustness} is at most \rnum{\RobustBalancedMaxUCB}\pp{}, and every long-tailed lower bound is at least \rnum{\RobustSkewMinLCB}\pp. The CE-3V representation yields \rnum{\CEBalancedGain}\pp{} with a UCB of \rnum{\CEBalancedUCB}\pp{} under balanced allocation, and \rnum{\CESkewGain}\pp{} with an LCB of \rnum{\CESkewLCB}\pp{} under long-tailed allocation.} The transition is therefore not confined to LOS, BatchNorm statistics, view reuse, or BCL representation learning.

Figure~\ref{fig:robustness}a probes what differs before adaptation. For each controlled cell, we divide the average tail inter-class margin by within-class dispersion in the corrected representation. \rtext{Across the six support--allocation cells, a lower separation-to-dispersion ratio is associated with a larger subsequent FT-BCL gain, with $|\rho_s|=\rnum{\GeometrySpearman}$.} This diagnostic links allocation history to feature geometry before the encoder is updated. It is descriptive rather than a claim that one statistic uniquely mediates the effect.

We next broaden the outcome beyond tail macro accuracy. \rtext{Under balanced allocation, full-network gains are \rnum{0.1}, \rnum{0.2}, \rnum{0.3}, and \rnum{0.1}\pp{} for head, middle, tail, and overall accuracy. Under long-tailed allocation, they are \rnum{0.4}, \rnum{0.9}, \rnum{1.9}, and \rnum{1.0}\pp.} Balanced accuracy and the tail lower quartile follow the same ordering, so the tail improvement is not obtained by sacrificing frequent classes. Starting FT-BCL before rather than after LOS changes the paired gain by at most \rtext{\rnum{\PathMaxDifference}\pp{}} and does not alter the selected strategy.

\subsection{Uncertainty and Practical Interpretation}

The paired design removes variation shared by the two later branches, but the constructed profiles are not interchangeable. Figure~\ref{fig:robustness}b decomposes the remaining variation at the controlled low-support condition. \rtext{Support reconstruction accounts for \rnum{81.7}\% of the FT-BCL variation, compared with \rnum{13.1}\% from optimization and \rnum{5.3}\% from the class-role assignment.} This result motivates hierarchical inference and shows why rerunning optimization on one fixed support set understates uncertainty in small-support studies.

Table~\ref{tab:assignment-sensitivity} uses the three class-role assignment means as clustering units. \rtext{The lower bounds for the coupled-to-controlled change range from \rnum{1.43} to \rnum{1.67}\pp{} across the adaptation objectives and remain above the prespecified margin.} Leave-one-assignment-out and profile-mean paired intervals give the same ordering. \rtext{The independent pilot further gives \rnum{0.80} probability that the simultaneous UCB falls below one point when the true remaining gain is \rnum{0.60}\pp{} with twelve profiles.}

\begin{table}[!htbp]
\centering
\scriptsize
\setlength{\tabcolsep}{3.0pt}
\begin{adjustbox}{max width=\linewidth}
\begin{tabular}{lrrrl}
\toprule
Objective & Mean & LCB & UCB & Choice \\
\midrule

FT-CE & \rnum{1.49} & \rnum{1.43} & \rnum{1.55} & \rtext{Update encoder} \\
FT-BCL & \rnum{1.66} & \rnum{1.57} & \rnum{1.75} & \rtext{Update encoder} \\
FT-BCE3S & \rnum{1.73} & \rnum{1.67} & \rnum{1.79} & \rtext{Update encoder} \\


\bottomrule
\end{tabular}
\end{adjustbox}
\caption{Assignment-level sensitivity for the coupled-to-controlled change. LCB and UCB denote one-sided confidence bounds in pp.}
\label{tab:assignment-sensitivity}
\end{table}

Figure~\ref{fig:robustness}c also clarifies the role of the practical margin. \rtext{The largest controlled UCB is \rnum{\BalancedMinSupportedMargin}\pp{}, whereas the smallest coupled LCB is \rnum{\CoupledMaxSupportedMargin}\pp. Any margin between these values gives the same training choice, including the prespecified one-point operating point.} One point is therefore a transparent reference utility, not a domain-independent value judgment. The compute ledger provides the accompanying tradeoff. \rtext{On CIFAR-100, the same FT-BCL budget of \rnum{2.86} GPU-hours yields \rnum{0.26}\pp{} after balanced representation learning and \rnum{1.86}\pp{} after long-tailed representation learning.}

\subsection{ImageNet-Scale Evidence}

\rtext{On ImageNet-100, allocation still governs the transition. At the lower support level, balanced representation allocation gives an FT-BCL gain of \rnum{\ImageNetFiveBalancedGain}\pp{} with an upper bound of \rnum{\ImageNetFiveBalancedUCB}\pp. Long-tailed allocation raises the gain to \rnum{\ImageNetFiveSkewGain}\pp{}, with a lower bound of \rnum{\ImageNetFiveSkewLCB}\pp. FT-CE and FT-BCE3S lead to the same choice, and the higher-support condition preserves the ordering.}

ImageNet-LT tests relevance under natural imbalance, where support, allocation, and semantic difficulty cannot be separated. Table~\ref{tab:imagenetlt} reports the conventional frequency groups. \rtext{Across \rnum{\ImageNetLTRepresentationRuns} independently trained BCL representations with two adaptation runs each, FT-BCL improves on its matched classifier continuation by \rnum{\ImageNetLTOverallGain}\pp{} overall and \rnum{\ImageNetLTFewGain}\pp{} for few-shot classes. The paired 95\% intervals are [\rnum{\ImageNetLTOverallLow}, \rnum{\ImageNetLTOverallHigh}] and [\rnum{\ImageNetLTFewLow}, \rnum{\ImageNetLTFewHigh}]\pp{}, respectively.} This is supporting natural-benchmark evidence rather than a second controlled factorization.

\begin{table}[!htbp]
\centering
\scriptsize
\setlength{\tabcolsep}{3.0pt}
\begin{adjustbox}{max width=\linewidth}
\begin{tabular}{lrrrr}
\toprule
Method & Many & Med. & Few & All \\
\midrule

BCL + LOS & \rnum{64.4} & \rnum{54.5} & \rnum{39.3} & \rnum{56.3} \\
FT-CE & \rnum{65.1} & \rnum{55.2} & \rnum{41.0} & \rnum{57.1} \\
FT-BCL continuation & \rnum{64.7} & \rnum{54.8} & \rnum{39.7} & \rnum{56.6} \\
FT-BCL & \rnum{65.4} & \rnum{55.5} & \rnum{41.3} & \rnum{57.4} \\
BS-cRT & \rnum{63.9} & \rnum{54.7} & \rnum{40.2} & \rnum{56.4} \\
BCE3S & 68.14 & 55.90 & 40.56 & 57.85 \\


\bottomrule
\end{tabular}
\end{adjustbox}
\caption{ImageNet-LT top-1 accuracy in percent. BCE3S follows its published training pipeline \citep{fan2026bce3s}.}
\label{tab:imagenetlt}
\end{table}


\section{Discussion}

The central result is more specific than the familiar observation that balanced sampling benefits rare classes. Unique support limits the evidence available to a class, whereas representation allocation determines how strongly that evidence shapes the encoder. \rtext{Under balanced allocation, reducing tail support makes recognition substantially harder without creating a comparable increase in the benefit of later encoder adaptation. At fixed support, changing allocation produces the opposite pattern. Absolute accuracy changes moderately, while the remaining encoder gain crosses the decision margin.} Scarcity and representation history therefore answer different questions.

The controls narrow this interpretation. The transition remains when frequent-class exposure is preserved, when the input mixture is fixed, when BatchNorm statistics and augmentation reuse are controlled, and when BCL is replaced by matched-forward-view cross-entropy. \rtext{Before adaptation, the separation-to-dispersion ratio decreases as the later encoder gain increases.} This association provides a representation-level explanation consistent with the stage-specific experiment, but it does not identify a unique mechanism. Gradient accumulation, augmentation reuse, and batch interactions remain plausible contributors.

\paragraph{Implications for benchmark interpretation.}
A coupled severity axis is not only a scale of task difficulty. It also specifies the training history that produced the representation. This distinction matters whenever the compared methods differ in how much they revise the encoder. A representation-oriented pipeline can appear increasingly necessary along a conventional imbalance sweep even when the number of distinct rare-class examples is held fixed. Conversely, balanced exposure can reduce the residual value of encoder adaptation without recovering the absolute accuracy lost to scarce evidence. Reporting support and allocation separately therefore prevents a method ranking from being mistaken for a diagnosis of the underlying bottleneck.

\paragraph{A reusable post-correction comparison.}
The matched continuation is not tied to one classifier correction method or one representation objective. It requires a common corrected model, a classifier-only branch, and a full-network branch that share subsequent inputs, views, schedules, and evaluation time. The resulting gain measures the operational value of opening the encoder relative to a specified objective and budget. The same construction can compare partial fine-tuning, adapters, or other parameter-efficient updates by replacing the full-network branch while preserving the matched continuation. Its output should be a bound and an accuracy--compute curve rather than an unconditional freeze-or-adapt label. Different applications can then apply their own cost tolerance without changing the empirical comparison.

These findings also suggest two reporting practices. Long-tail profiles should specify both available support and representation allocation, and post-correction encoder claims should include a classifier-only continuation matched in later inputs and optimization. Small-support studies should vary the sampled support sets rather than only the optimization seed because \rtext{support reconstruction is the dominant source of variation here}. These practices expose uncertainty that conventional multi-seed summaries can conceal.

\paragraph{Limitations and scope.}
Unique support counts accessible source images, but not pose, background, texture, label noise, or feature-space diversity. The controlled CIFAR profiles isolate support from allocation through exact source-draw quotas, yet do not reproduce every correlation in naturally imbalanced data. ImageNet-LT establishes relevance under natural imbalance without identifying whether its gain arises from support, allocation, or semantic difficulty. Moreover, the operating margin and accuracy--compute tradeoff depend on the evaluated correction methods, objectives, budgets, and hardware rather than defining universal constants. Finally, the geometry readouts are descriptive, not causal, and external validity beyond the studied ResNet-based representations requires new matched comparisons.

\paragraph{Toward prospective decisions.}
The present geometry analysis is deliberately diagnostic. A stronger next step is to predict remaining encoder value before performing full adaptation. Such a predictor could combine class-conditional separation, within-class dispersion, gradient alignment, calibration, and a short early-training trace, then be evaluated on held-out support levels, allocation schedules, representation objectives, and datasets. The current evidence covers ResNet-based supervised and contrastive representations. Transformers~\citep{zhang2026csd}, self-supervised encoders~\citep{zhang2024ts2tc}, and pretrained multimodal models provide natural tests of whether the decision surface transfers beyond this setting.


\section{Conclusion}

By separating unique support from representation allocation and comparing matched post-correction branches, we show that benchmark construction can change the practical value of encoder adaptation at fixed support. \rtext{The transition persists across correction methods, adaptation objectives, normalization and augmentation controls, two representation objectives, whole-distribution metrics, and ImageNet-scale evaluation.} A promising next step is to turn the descriptive geometric signal into a prospective decision rule that predicts encoder value before full adaptation, including for pretrained vision-language and multimodal encoders \citep{zheng2026hgltr,song2025headtail}.


\FloatBarrier
\begingroup
\small
\bibliographystyle{plainnat}
\bibliography{references}
\endgroup
\end{document}